\documentclass[reprint]{JASA}

\usepackage{multirow}
\usepackage{makecell}

\begin{document}

\title[How Does a Deep Neural Network Look at Lexical Stress in English Words?]{How Does a Deep Neural Network Look at Lexical Stress in English Words?}

\author{Itai Allouche}
\author{Itay Asael}
\author{Rotem Rousso}
\author{Vered Dassa}
\affiliation{Faculty of Electrical and Computer Engineering,  Technion -- Israel Institute of Technology, Haifa, 3200003, Israel}
\author{Ann Bradlow}
\author{Seung-Eun Kim}
\author{Matthew Goldrick}
\affiliation{Department of Linguistics, Northwestern University, Evanston, Illinois 60208, USA}
\author{Joseph Keshet}
\email{jkeshet@technion.ac.il}
\affiliation{Faculty of Electrical and Computer Engineering, Technion -- Israel Institute of Technology, Haifa, 3200003, Israel}



\begin{abstract}
Despite their success in speech processing, neural networks often operate as black boxes, prompting the question: what informs their decisions, and how can we interpret them? This work examines this issue in the context of lexical stress. A dataset of English disyllabic words was automatically constructed from read and spontaneous speech. Several Convolutional Neural Network (CNN) architectures were trained to predict stress position from a spectrographic representation of disyllabic words lacking minimal stress pairs  (e.g., initial stress \emph{WAllet}, final stress \emph{exTEND}), achieving up to 92\% accuracy on held-out test data. Layerwise Relevance Propagation (LRP), a technique for neural network interpretability analysis, revealed that predictions for held-out minimal pairs (\emph{PROtest} vs. \emph{proTEST}) were most strongly influenced by information in stressed versus unstressed syllables, particularly the spectral properties of stressed vowels. However, the classifiers also attended to information throughout the word. A feature-specific relevance analysis is proposed, and its results suggest that our best-performing classifier is strongly influenced by the stressed vowel's first and second formants, with some evidence that its pitch and third formant also contribute. These results reveal deep learning's ability to acquire distributed cues to stress from naturally occurring data, extending traditional phonetic work based around highly controlled stimuli. 
\end{abstract}


\maketitle

\section{\label{sec:1} Introduction}

While deep learning based architectures have transformed many aspects of speech processing, those models often exhibit an nontransparent prediction process, obscuring how specific decisions are made. This work explores the application of one technique for shedding light on such architectures through the lens of a well-studied phenomenon in speech science and speech processing: lexical stress. Lexical stress refers to the prominence, accentuation, or emphasis of a particular syllable within a word. In many languages, lexical stress is crucial for communication as it can serve as the primary basis for a contrast in meaning. For example, in English, the words \emph{''PROtest''} (noun; stress on initial syllable) and \emph{''proTEST''} (verb; stress on final) have contrasting stress patterns but the same phonemes (with predictable changes to the quality of the initial vowel depending on stress).

A long tradition of phonetics research has examined the (relative) temporal, spectral, and energetic properties of stressed and unstressed syllables \cite{fry1955duration, fry1958experiments}; for reviews, see \citet{cutler}, \citet{gay1978physiological}, \citet{gordon2017acoustic}, and \citet{vanHeuvenStress}. Building on this work, automatic stress detection methods have  used acoustic features such as fundamental frequency, amplitude, and energy \cite{lieberman1960some, 1415269}. Bayesian classifiers with multivariate Gaussian distributions have been applied to enhance stress detection accuracy, utilizing acoustic features such as peak-to-peak amplitude, normalized energy, and vowel duration \cite{1168788, 607932}. Similar to other areas of speech processing, recent work using deep learning architectures has significantly improved over these more traditional approaches. Convolutional Neural Networks (CNNs) and Multi-Distribution Deep Neural Networks (MD-DNNs) outperformed traditional models in both English and Arabic stress detection in second language speech \cite{shahin2016automatic, LI201828}. An attention-based neural network combined with data augmentation through a neural network based text-to-speech system has been introduced to detect lexical stress errors in second-language English learners, achieving significant improvements in precision and recall \cite{korzekwa2020detection}. Classifiers applied to self-supervised learning representations from wav2vec 2.0 rather than traditional acoustic features also show high performance \cite{bentum2024processing}.

Although these advancements in deep learning have led to significant improvements in speech processing, they also introduce challenges, particularly in understanding classifier decisions and gaining insight into how these models operate. 
This issue of interpretability is crucial. Better understanding the acoustic features that drive model predictions can help us verify that the model will be reliable when applied to novel data (which may have different distributions for critical features). Neural network interpretability has broadly become an active research area, with a number of techniques being explored across different domains. Here, we focus on Layerwise Relevance Propagation (LRP) \cite{bach2015pixel}, a technique that was developed to provide insights into the decision-making processes of these complex models by understanding what specific portions of the input influence the model's output. We use this technique to investigate how CNNs -- which have demonstrated superior performance in lexical stress detection -- derive their predictions from the spectral features in their input. 

This paper offers several novel contributions. First, a new speech dataset was automatically gathered without human annotation, using a pipeline built around deep learning systems. Second, we present a state-of-the-art model for lexical stress classification in disyllabic words and show its ability to generalize to novel disyllabic words. Lastly, we employ LRP to analyze the spectral features these classifiers use to achieve high accuracy.

This paper is organized as follows. In the next section, we describe how the dataset was generated. In Section~\ref{sec:3} we present how we built the classifier and how it was trained. In Section~\ref{sec:4} we present the LRP technique, its application in our study, and our techniques for analyzing LRP results. In Section \ref{sec:5} we empirically evaluate the classifiers. The LRP analyses are presented in Section~\ref{sec:6}, and the paper concludes with a discussion in Section~\ref{sec:7}.

\begin{figure*}[htbp]
\centering
\includegraphics[width=\linewidth]{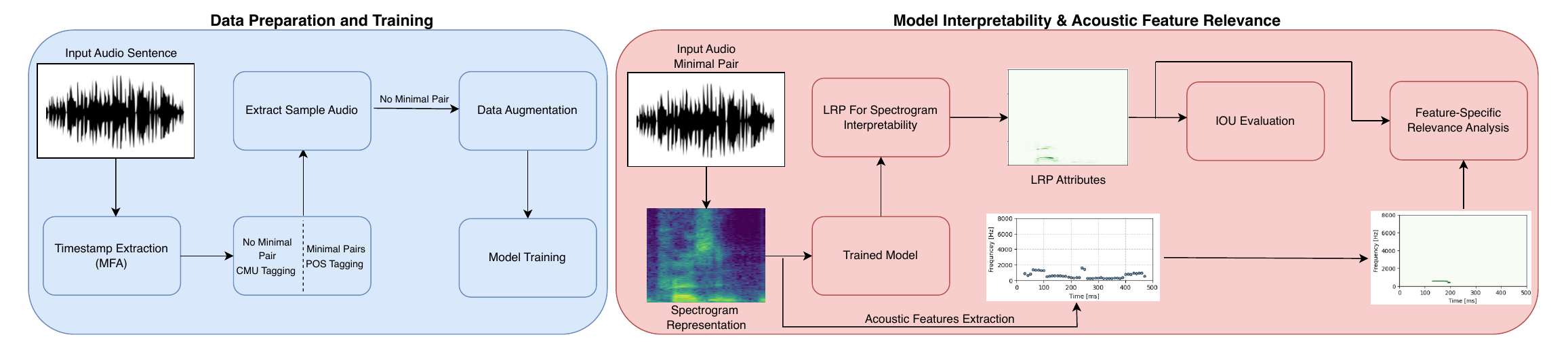}
\caption{End-to-end workflow for dataset construction, model training, and interpretability analysis. The blue-colored section illustrates the dataset creation and training process. Audio is collected from various datasets and extracted based on word- and phoneme-level timestamps. Words are tagged for stress using either part-of-speech tagging (minimal pairs) or dictionary entries (no minimal pair words). Once the audio is extracted, the training data (no minimal pair words) is expanded through data augmentation techniques to make CNN-based model training more robust. The red-colored section illustrates the processes of model interpretability and acoustic feature relevance measurement, applied to the test set of minimal pairs. Audio is converted into a spectrogram. This is processed by the trained model; LRP methods are applied to generate heatmaps, revealing the regions of the spectrograms that contributed most to the model’s predictions. The contribution of different (sub)syllabic regions to the LRP heatmap is quantified with IOU metric. For the stressed vowel, acoustic features are extracted and used to generate feature-specific heatmaps; these are compared to the full heatmap to determine which acoustic features are most dominant in the model’s decision-making process.}
\label{fig:fig1}
\end{figure*}

\section{Dataset Construction}\label{sec:2}

Previous studies on lexical stress often depended on stress labels assigned by human annotators \cite{shahin2016automatic, korzekwa2020detection, 1415269} or were assigned based on the canonical stress for the orthographic word taken from pronunciation dictionaries such as the CMU dictionary \cite{LI201828}. However, manual annotation is costly and can introduce inter-annotator inconsistency, and the use of a single citation form can yield errors for orthographically ambiguous minimal pair words (e.g., protest). Prior work also failed to explicitly document measures to prevent lexical overlap between training and evaluation splits, leaving open the possibility that identical word types were seen during both stages \cite{shahin2016automatic, korzekwa2020detection, 1415269, LI201828}. Reported accuracy may therefore overestimate true generalization. To avoid these concerns, we created a novel dataset, using a fully automatic pipeline, without utilizing human annotations \citep{mlspeech_lexical_stress_2025}.

\subsection{Selection of English Disyllabic Words}\label{subsec:2:1}

We prompted ChatGPT 4o \cite{hurst2024gpt} to generate 30 pairs of English disyllabic words that form stress minimal pairs: pairs of words with the same spelling but different meaning depending on the primary stress placement. Five of these items could not be incorporated into other parts of our pipeline (the part-of-speech tagger, discussed below, failed to correctly annotate these forms). The remaining 25 pairs are \emph{address}, \emph{conduct}, \emph{content}, \emph{contrast}, \emph{convert}, \emph{decrease}, \emph{digest}, \emph{export}, \emph{extract}, \emph{import}, \emph{increase}, \emph{insult}, \emph{object},  \emph{perfect}, \emph{present}, \emph{progress}, \emph{project}, \emph{protest}, \emph{rebel}, \emph{record}, \emph{refuse}, \emph{reject}, \emph{subject}, \emph{suspect}, and \emph{transfer}; the initial and final syllable stress variants of these words will be referred to as ''minimal pairs.'' We also asked ChatGPT 4o to generate a larger set of 250 English disyllabic words that do not have stress minimal pairs (e.g., initial stress \emph{WALLet}, final stress \emph{exTEND}). After excluding one error, we retained 249 words (124 initial stress, 125 final stress). We refer to this as the set of  ''no minimal pair words.''

\subsection{Extraction of Words from Existing Corpora}\label{subsec:2:2}

These words were then extracted from three existing English speech corpora. The first dataset is LibriSpeech \cite{7178964}, a widely-used corpus consisting of approximately 1,000 hours of read  speech derived from audiobooks. The second dataset is The Supreme Court corpus \cite{spaeth2014supreme}, which includes recordings of oral arguments made before the United States Supreme Court. This dataset provides speech data from a highly formal and legal context, offering variation in speaker formality and legal terminology that can influence acoustics associated with lexical stress. The third dataset is TED-LIUM \cite{hernandez2018ted}, derived from TED Talks. It includes over 400 hours of speech by speakers from various language backgrounds and nationalities, discussing a wide range of topics. The diversity in speaking styles and contexts in TED-LIUM further enriches the phonetic variation in our dataset.

The processing pipeline is shown in Figure \ref{fig:fig1}. First, we used a forced alignment algorithm to align the transcript with the audio. Specifically we used Montreal Forced Aligner (MFA) \cite{mcauliffe2017montreal} to align LibriSpeech and Supreme Court. MFA works at time-resolution of 10 millisecond and shown by \citet{rousso2024tradition} to outperform recent end-to-end ASR-based aligners such as WhisperX \cite{bain2023whisperx} and MMS \cite{pratap2024scaling}. For TED-LIUM, which already includes word-level timestamps, we used the provided alignments. Importantly, unlike MFA-aligned corpora (LibriSpeech and Supreme Court), where word boundaries are adjacent (e.g., ''hello 0–0.3 second'', ''my 0.3–0.4''), the TED-LIUM timestamps may be non-adjacent (e.g., ''hello 0–0.3'', ''my 0.35–0.4''), leaving short pauses between words.

Once the start time of each word is obtained, a 0.5-second window is opened. This window size was chosen to ensure that both syllables of every disyllabic word were fully included and to maintain a uniform input length compatible with our CNN architecture. We retained only those whose total duration was less than 0.5 second and zero-padded them as needed to reach exactly 0.5 second. Each minimal pair word was further segmented into syllables using a syllabification algorithm \citep{p2tk_syllabifier_2014}. In addition, phoneme-level timestamps derived from the forced aligner were incorporated.

After identifying the location of target minimal pair words within the audio, we applied part-of-speech tagging system \cite[SpaCy;][]{Annex2020} to the entire audio in order to determine whether each word was used as noun or verb. The tagging follows linguistic conventions for English disyllabic words, where nouns typically carry primary stress on the initial syllable, referred to as initial stress (IS), and verbs carry primary stress on the final syllable, referred to as final stress (FS) \cite{lieberman1960some}. For the no minimal pair, we used annotations from the CMU Pronouncing Dictionary \cite{LI201828}, as the stress placement is unambiguous in this case. During this process, we observed that certain words, specifically \emph{address}, \emph{content}, \emph{export}, \emph{extract}, and \emph{decrease}, were particularly affected by inaccuracies in the part-of-speech tagging system. Many of these errors involved incorrectly labeling words with initial stress instead of final stress. To correct these inconsistencies, we manually reviewed and relabeled all instances of these words.

Overall, we had 7,446 samples (124 types) and 3,263 samples (125 types) for initial and final stress,  respectively, in the no minimal pairs set, and 5,475 samples and 1,715 samples for initial and final stress, respectively, in the minimal pairs set, consisting of a total of 25 pairs.

\subsection{Phonetic Properties of Extracted Samples}\label{subsec:2:3}

To provide an initial verification that our procedures yielded sets of disyllabic words that contrasted in stress location, two well-known correlates of lexical stress were examined: amplitude (vowels in stressed syllables are louder than those in unstressed syllables) and duration (vowels in stressed syllables are longer than those in unstressed syllables) \cite{fry1955duration,fry1958experiments}. For each sample, we calculated the ratio of the properties of the vowels of the initial versus final syllables. These relative measures allow us to control for potential differences in speech rate and amplitude across samples. Our samples reliably showed the expected differences. Samples with initial stress had high ratios of amplitude (mean 0.6, s.d. 0.18) and duration (mean 0.53, s.d. 0.15), reflecting relatively loud, long vowels in stressed initial versus unstressed final syllables. In contrast, samples with final stress showed lower mean ratios for amplitude (mean 0.29, s.d. 0.14) and duration (mean 0.35, s.d. 0.12), reflecting relatively quiet, short vowels in unstressed initial versus stressed final syllables. To confirm that this is a reliable difference between sample types, we used bootstrap resampling (with 1,000 replicates) to estimate the 95\% confidence interval for differences in mean ratios between initial versus final stress samples. These differences were significant (amplitude mean difference: 0.305; 95\% CI (0.302,0.309); $p<$.0001; duration mean difference: 0.178, 95\% CI (0.175,0.18), $p <$ .0001). This provides independent confirmation that the samples in this data set have contrasting stress patterns.

\begin{figure*}[htbp]
\centering
\includegraphics[width=\linewidth]{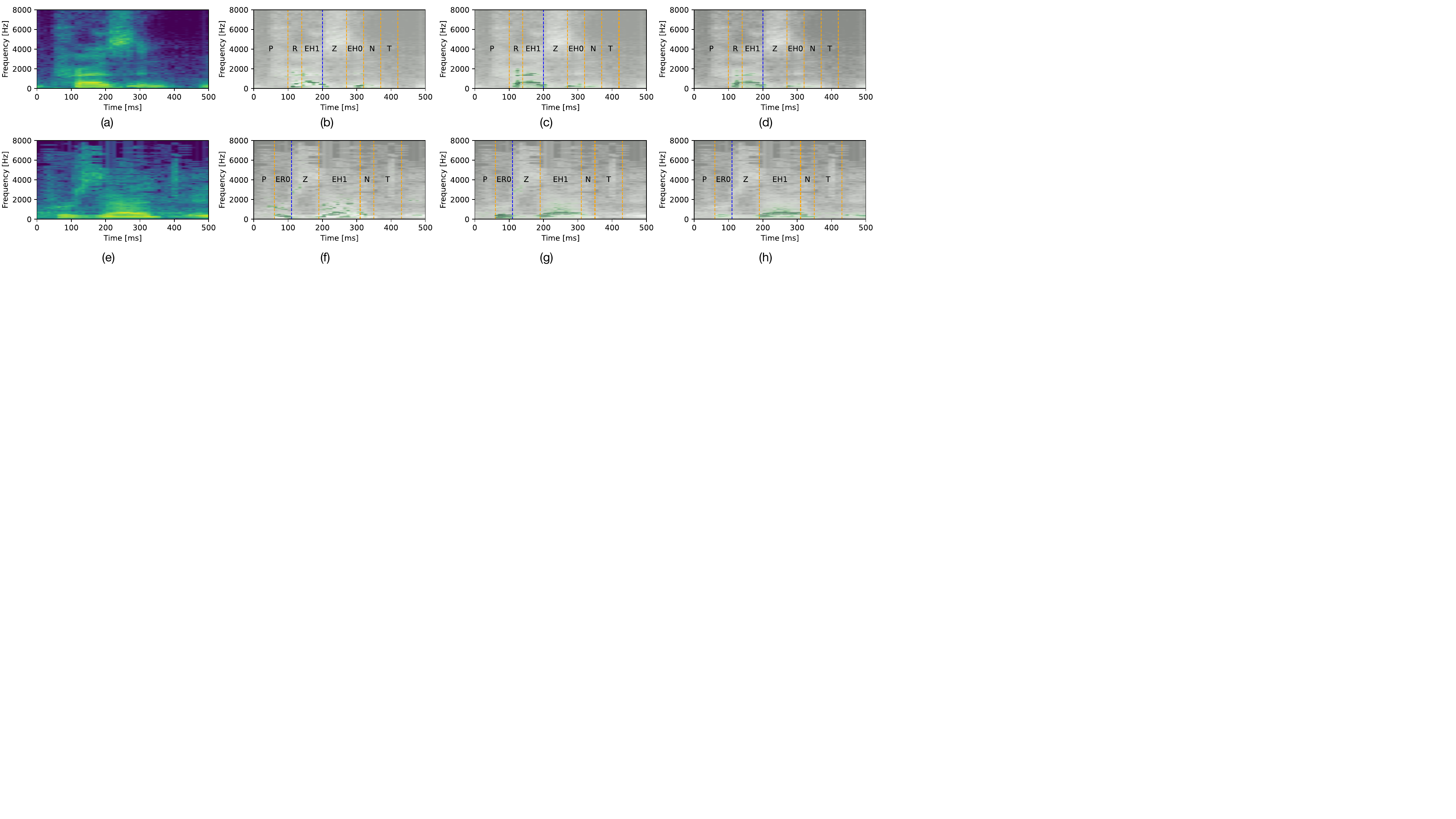}
    \caption{Spectrograms and corresponding attributions for the words \emph{''PREsent''} and \emph{''preSENT''} using the VGG16 model. Orange vertical lines represent phoneme boundaries,
    with each label centered within its phoneme segment; 1 and 0/2 denote stressed and unstressed vowels, respectively. The blue vertical line represents the end of the initial syllable. Panel (a) shows the spectrogram of \emph{''PREsent''} (IS), while panel (e) shows the spectrogram of \emph{''preSENT''} (FS). Panels (b)-(d) and (f)-(h) display attributions using various methods:  $\mathbf{LRP_{\epsilon}}, \mathbf{LRP_{\alpha1}} \text{ and } \mathbf{LRP_{CMP}}$ respectively, shown over the spectrograms. These LRP variants effectively capture initial versus final stress contrasts, highlighting features at different scales. The $\mathbf{LRP_{\alpha1}}$ (panel (c) vs. (g))and $\mathbf{LRP_{CMP}}$ methods (panel (d) vs. (h)) show clearly different patterns depending on the location of stress. Note that the phoneme labels differ slightly across the two rows; these reflect natural differences in pronunciation between initial and final stress members of this word pair (i.e., stress-related changes to vowel quality).}
    \label{fig:fig2}
\end{figure*}

\section{Lexical Stress Detection}\label{sec:3}

\subsection{Model Architectures}\label{subsec:3:1}

In our study, we investigated different CNN architectures to develop a binary classifier for identifying the initial or final stress of stress-minimal word pairs. The input to all models consists of magnitude spectrograms obtained using the Short-Time Fourier Transform (STFT), following z-score normalization (see Figures \ref{fig:fig2} and \ref{fig:fig3} for examples). The STFT was computed with a sampling rate of 16,000 Hz, using a Hamming window of size 0.02 seconds and a window stride of 0.01 seconds. We experimented with several well-known CNN architectures. The LeNet-5 model consists of 3 connected convolutional layers with average pooling between the layers, followed by 2 dense layers in a 2x2 configuration \cite{726791}. The VGG11, VGG16, and VGG19 architectures consist of 8, 13, and 16 convolutional layers, respectively, with max pooling between the layers, followed by an additional 3 dense layers \cite{simonyan2014very}. Finally, the ResNet18 model consists of 18 convolutional layers with residual connections to address the vanishing gradient problem during training \cite{He2016CVPR}.

\subsection{\label{subsec:3:2} Data Splitting}

The data were partitioned so that no word type appears in more than one split, thereby eliminating lexical overlap between training, validation, and test sets.

The training set was constructed using 201 out of the 249 word types from the no minimal pair words. Data augmentation techniques were employed to enlarge this set of examples. For each sample, we added a low-pass filtered version of the sample (cutoff frequency at 3000 Hz) as well as three versions of the sample mixed with noise. A sample of multi-talker babble from the VOiCES Corpus \cite{richey2018voices} was combined with the original sample at three different levels (20 dB, 10 dB, and 3 dB SNR). These augmentations aimed to simulate real-world conditions, enhancing the robustness of our models. After the augmentation, we had 29,830 and 13,510 samples from the initial and final stress, respectively, corresponding to the 201 no minimal pair word types. The augmented dataset will be referred to as the ''no minimal pairs train.''

The remaining 48 no minimal pair word types were divided into two lists of 24 words each, with one list used for the validation (i.e., hyper-parameter tuning; ''no minimal pairs validation'') and the other for the test set (''no minimal pairs test''). The 25 minimal pairs words will be referred to as ''minimal pairs test''. Note that two test sets were used (''no minimal pairs test'' and ''minimal pairs test'') as the minimal pairs set may contain errors; stress was determined solely by part-of-speech tagging, which may be inaccurate. In contrast, the tagging for the no minimal pair words is more reliable; as the word form is unambiguous, the lexically-determined stress pattern can be accurately identified. See Table \ref{table:tab1} for dataset composition.

\begin{table}[ht]
    \centering
    \begin{ruledtabular}
    \caption{Composition of the  train (before and after data augmentations), validation, and test sets. Each set consists of disjoint word types (no overlap across splits). The table reports the number of unique word types and the total number of samples corresponding to initial stress (IS) and final stress (FS) words.}
    \label{table:tab1}
    \begin{tabular}{l|ccc}
        Set & \shortstack{\#Word\\types} & \shortstack{\#IS\\samples\\(w. aug.}) & \shortstack{\#FS\\samples\\(w. aug.}) \\
        \hline
        \noalign{\vskip 2pt}
        \raggedright no minimal pairs train & 201 & \shortstack{5966\\ (29,830)}& \shortstack{2702\\ (13,510)} \\
        \raggedright no minimal pairs validation & 24 & 763 & 260 \\
        \raggedright no minimal pairs test & 24 & 717 & 301 \\
        \raggedright minimal pairs test & 25 & 5475 & 1715 \\
    \end{tabular}
    \end{ruledtabular}
\end{table}

\subsection{\label{subsec:3:3} Training Process}
The models were trained using standard procedures for CNNs. To stabilize training, we employed an Adam Optimizer \cite{kingma2014adam} with a learning rate $5\times10^{-4}$ and batch normalization. Training was performed for up to 50 epochs, with early stopping triggered after 15 epochs without validation improvement. All experiments were conducted on 8 NVIDIA A40 GPUs.

As the training data contain substantially more initial than final stress examples, a potential concern is an initial stress bias, whereby a classifier could achieve relatively high accuracy by disproportionately predicting the majority class, rather than relying on acoustically relevant cues to lexical stress. Such a bias could obscure the extent to which the model genuinely learns stress-related acoustic patterns.
To mitigate this possibility, we employed the \emph{focal loss} function \cite{lin2017focal}, modifying the standard cross-entropy loss to focus more on hard-to-classify examples and to reduce the influence of the majority class during training. Focal loss (\emph{FL}) introduces a modulating factor $(1-p_t)^\gamma$ to the cross-entropy objective, reducing the relative loss contribution of well-classified examples and amplifying that of misclassified ones. Formally, it is defined as:
\begin{equation}
    FL(p_t)=-(1-p_t)^{\gamma}\log(p_t)~,
\end{equation}
where $p_t$ denotes the predicted probability for the true class, and $\gamma \geq0$ is the focusing parameter that controls the rate at which easy examples are down-weighted. In our experiments, we set $\gamma=2$ based on validation performance, which provided a good balance between stability and discrimination.

\section{\label{sec:4}Interpretability Analysis Methods}

\subsection{\label{subsec:4:1} Layer-wise Relevance Propagation}

Layer-wise Relevance Propagation (LRP) has been extensively used in computer vision to provide insights into the contribution of individual input features to model predictions \cite{9206975}. Starting at the output, LRP attributes the classification prediction to units in the preceding layer -- how \emph{relevant} is each unit to the classifier producing one output versus another? This process is then repeated, back-propagating the relevance score of each unit to the classification prediction until the input layer is reached. This technique allows us to attribute relevance scores to each feature (e.g., how much does pixel X contribute to classifying an image as \emph{dog} versus \emph{cat}?). 

In our study, we apply this technique to our lexical stress classifiers, backpropagating prediction to time-frequency bins in the input using the Captum interpretability library \citep{kokhlikyan_captum_2020}.
This method identifies the regions of the spectrogram that have the most significant influence on the model’s classification of lexical stress. We specifically examine our best-performing VGG-based classifier, VGG16, and ResNet18 (for model performance comparisons see Section \ref{sec:5} and Table \ref{table:tab2}).

\begin{figure*}[htbp]
\centering
\includegraphics[width=\linewidth]{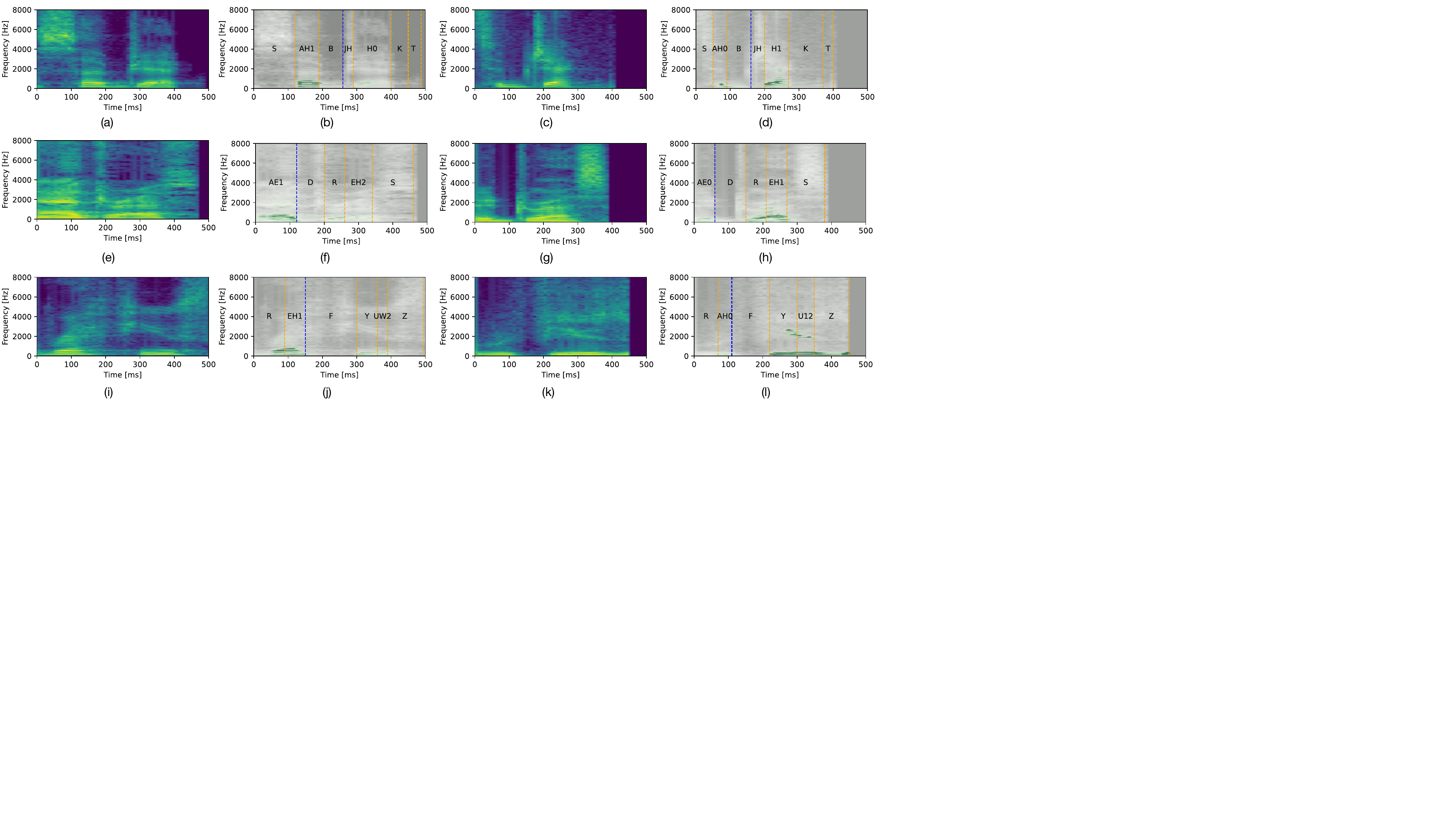}
    \caption{Further illustrative examples of spectrograms and corresponding attributions for the words ''\emph{subject},'' ''\emph{address}'' and ''\emph{refuse}'' with different primary stress location, using the VGG16 model. Vertical lines and annotation follow Figure \ref{fig:fig2}. Panel (a)-(d) show spectrograms and corresponding  \begin{math} \mathbf{LRP_{CMP}}\end{math} heatmaps for the word "\emph{subject}" with initial and final stress, respectively. Similarly, panels (e)-(l) display spectrograms and heatmaps of the words ''\emph{address}'' and ''\emph{refuse},'' with different stress types. Note that the examples were drawn from datasets in which word segments were zero-padded to reach the fixed 0.5 second window length, as described in Section \ref{subsec:2:2}.}
    \label{fig:fig3}
\end{figure*}

Denote by $\mathbf{x}^l$ the input to the $l$-th layer. Denote by $\mathbf{W}^l$ the weights associated with layer $l$. Denote by $\mathbf{Z}^l$ is the elementwise preactivation of layer $l$ and it is defined as $Z^l_{ij}={W}^l_{ij} \odot {x}^l_{i}$. The relevance score of layer $l$ is computed recursively from a higher layer $l+1$. In \citet{9206975} the relevance score $\mathbf{LRP_Z}$ is defined as 
\begin{equation}
R_i^{l}=\sum_j \frac{Z^l_{ij}}{Z^l_j} R_j^{l+1}~,
\end{equation}
where $Z^l_j=\sum_i Z^l_{ij}$.
The  \begin{math}\mathbf{LRP}_Z\end{math} method effectively traces the influence of each element of the input through the network, providing insight into how these impact the model's predictions.

In addition to this method, we experimented with several types of LRP to determine the most effective method for identifying the regions of the spectrogram that are most influential for classifying words as stressed versus unstressed. The $\mathbf{LRP}_\epsilon$ rule extends the basic LRP method by addressing challenges such as division by zero and the influence of weak or noisy preactivations. It is defined as:
\begin{equation}
R_i^{l}=\sum_j \frac{Z^l_{ij}}{Z^l_j + \epsilon\,\mathrm{sign}(Z^l_j)} R_j^{l+1}~,
\end{equation}
In this rule, a small constant \begin{math}\epsilon\end{math} is added to the denominator along with the sign of \begin{math}Z^l_j\end{math}. This modification prevents division by zero and mitigates the impact of small or noisy preactivations \begin{math}Z^l_{ij}\end{math} on the relevance distribution. By incorporating \begin{math}\epsilon\end{math}, the \begin{math}\mathbf{LRP}_\epsilon\end{math} rule enhances the stability of relevance propagation, leading to more reliable and interpretable results. This adjustment is particularly beneficial when dealing with minimal or problematic contributions, as outlined by \citet{bach2015pixel}.

The $\mathbf{LRP_{\alpha\beta}}$ rule, introduced by  \citet{bach2015pixel}, is an extension of the basic LRP method that differentiates the contributions of facilitatory and inhibitory activation flow. It is defined as:
\begin{equation}
R_i^{l} = \sum_j \left( \frac{\alpha Z_{ij}^+}{Z_j^+} + \frac{\beta Z_{ij}^-}{Z_j^-} \right) R_j^{l+1}
\end{equation}
where $Z_{ij}^+ = \max\{Z^l_{ij}, 0\}$ and $Z_{ij}^- = \min\{Z^l_{ij}, 0\}$. In this rule, \begin{math}\alpha\end{math} and \begin{math}\beta\end{math} are non-negative parameters that weight the relevance distribution for facilitatory and inhibitory preactivations, respectively. The sum of \begin{math}\alpha\end{math} and \begin{math}\beta\end{math} is constrained to 1 to ensure the conservation of relevance between layers. This approach allows for a more nuanced allocation of relevance based on the nature of the preactivations.

Earlier implementations of LRP, such as those proposed by \citet{bach2015pixel} and \citet{Lapuschkin2016CVPR}, typically employed a single LRP rule applied uniformly throughout the network.  This approach failed to account for the varying needs of different layers or components within the network, resulting in less accurate and insightful relevance attributions. To address these limitations, \citet{9206975} introduced a composite strategy, which employs a combination of multiple LRP rules tailored to different parts of the network. In our analysis, an identity function was used to backpropagate activation for the first two convolutional layers closest to the input, $\mathbf{LRP}_{\alpha1}$ ($\mathbf{LRP}_{\alpha\beta}$ with $\alpha = 1$) for the remaining convolutional layers, and $\mathbf{LRP}{\epsilon}$ for the fully connected layers closest to the output. This combined approach, denoted as $\mathbf{LRP_{CMP}}$, enhances model interpretability and provides more accurate relevance attributions.

\subsection{\label{subsec:4:2} Intersection over Union in Sub-Syllabic Regions of the LRP heatmap}
We quantitatively assess the LRP analysis to assess how regions corresponding to different components of the initial and final syllables contribute to lexical stress classification using the Intersection Over Union (IOU) metric, adapted for analyzing relevance attribution in heatmaps. Specifically, we define the inside-total overlap ratio \begin{math}\mu\end{math}:
\begin{equation}
    \mu=\frac{R_{in}}{R_{tot}}
\end{equation}
Here, \begin{math}R_{in}\end{math} is the sum of all pixels of the spectrogram inside a bounding box derived from LRP heatmap, and \begin{math}R_{tot}\end{math}  is the total sum of relevance values across the entire heatmap. This metric \begin{math}\mu\end{math} effectively measures the proportion of relevance attributed to the area of interest, providing insight into what portions of the signal the model has paid attention to. Higher values of \begin{math}\mu\end{math} indicate that a larger fraction of the relevance is concentrated within the target area. 

The bounding boxes defining the regions of analysis were created using the phoneme-level timestamps and syllable boundaries described above, ensuring alignment with the relevant parts of the spectrograms. We set bounding boxes around the initial and final syllables and then, within each syllable, separate the vowel region from any other portion of the syllable (onset/coda).

\subsection{\label{subsec:4:3} Feature-Specific Relevance}

In addition to the IOU metric, we introduce a measure to evaluate how the model focuses on specific acoustic features during decision-making. Focusing on the vowel within the stressed syllable, this analysis aims to isolate the contribution of each acoustic feature to the model's decision-making process. We constructed \emph{feature-specific heatmaps} that represent the relevance distribution as if the model were focusing exclusively on the given feature. This was done by extracting the fundamental frequency ($F_0$), the first three formants ($F_1, F_2$ and $F_3$) along with their formant bandwidth and sound intensity using the default settings of the Praat analysis software \cite{pratt}. For each formant, heatmap values at each time point were generated by normalizing the intensity within the formant's bandwidth and then scaling these normalized values by the sound intensity. 

To quantify the similarity between these feature-specific heatmaps and the original relevance heatmap produced by LRP, we employed the Pearson Correlation Coefficient $r$. By comparing the strength of the correlation between feature-specific heatmaps, we gain insight into the relative importance of each acoustic feature in the model's decision-making process. Additionally, we examine the ability of heatmaps created by all possible combinations of these features to determine how these different acoustic features interact to contribute to the model's predictions.

Finally, we also examine the limitations of these specific features by considering the distribution of residual pixels left unexplained by any of these features ($F_0$ to $F_3$). We mask the LRP heatmap using the heatmap generated by combining all these features. We then examine the relative distribution of these unexplained pixels across the spectrum of the stressed vowels to consider what additional features, beyond pitch and formants, contribute to model performance.

\section{\label{sec:5} Model Performance}

Classifier results on the test sets are summarized in Table \ref{table:tab2}. The VGG16 architecture achieved classification accuracies of 88\% and 92\% in distinguishing between initial and final stress in disyllabic minimal pair and no minimal pair words test sets, respectively. The lower accuracy in the minimal pair test set may reflect errors in part-of-speech tagging. Note as well that human perception of lexical stress in disyllabic words based solely on acoustics (without access to syntactic or other contextual cues) is not at ceiling; \citet{yu2010cross} reported first-language English listeners had a mean accuracy of 84\% in this task. 

Other architectures also showed strong results. VGG19 matched VGG16's performance, despite utilizing a more complex architecture, suggesting that VGG16 is sufficient. VGG11 showed slightly lower performance on both test sets. While the other architectures also showed slightly lower performance, they still achieved a high level of performance ($>$ 86\% on both test sets).

\begin{table}[ht]
    \centering
    \begin{ruledtabular}
    \caption{Classification accuracy for each model architecture on each stress type and test set (Chance-level performance is 50\%).}
    \label{table:tab2}
    \begin{tabular}{l|l|ccc}
        \multirow{2}{*}{Test Set} & \multirow{2}{*}{Classifier} 
        & \multicolumn{3}{c}{Accuracy (\%)} \\
        & & IS & FS & Overall \\ 
        \hline
        \multirow{4}{*}{\vspace{-0.5cm}\centering \mbox{minimal pair}}
        & LeNet & 86.3 & 82.5 & 86\\
        & VGG11 & 87.2 & 83.5 & 87\\
        & VGG16 & 88.2 & 83.6 & \textbf{88}\\
        & VGG19 & 88.3 & 82.3 & \textbf{88}\\
        & ResNet18 & 87.74 & 84 & 87\\
        \hline
        \multirow{4}{*}{\vspace{-0.5cm}\centering \mbox{no minimal pair~}}
        & LeNet & 91 & 85.6 & 90\\
        & VGG11 & 91.8 & 88.3 & 91\\
        & VGG16 & 92.6 & 90.6 & \textbf{92}\\
        & VGG19 & 92.2 & 91.3 & \textbf{92}\\
        & ResNet18 & 90.3 & 88.7 & 90\\
    \end{tabular}
    \end{ruledtabular}
\end{table}

\section{\label{sec:6}Interpretability Results}

We utilized LRP and our interpretability metrics to examine more closely the aspects of the input that influenced the strong performance of these classifiers.

\subsection{\label{subsec:6:3} Layer-wise Relevance Propagation (LRP) Analysis}
Inspection of the LRP heatmaps for both initial and final stress words suggested they highlighted distinct areas of relevance that align with the energy distribution patterns observed in the spectrograms. Figures \ref{fig:fig2} and \ref{fig:fig3} illustrate this for the best-performing VGG16 classifier. For initial stress words, the LRP heatmaps emphasized the initial syllable as the most relevant area for classification by VGG16. 
Conversely, for final stress words, the LRP heatmaps showed a concentration of relevance towards the end of the phoneme sequence. Figure \ref{fig:fig3} provides further illustrations of these differences, focusing on the $\mathbf{LRP_{CMP}}$ method.

\begin{figure*}[htbp]
    \centering
    \figline{
    \leftfig{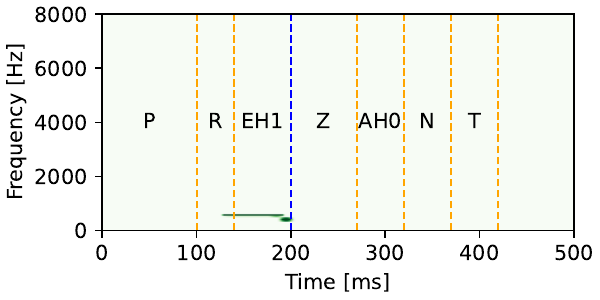}{6.0cm}{(a) F1 Heatmap}\hfill
    \leftfig{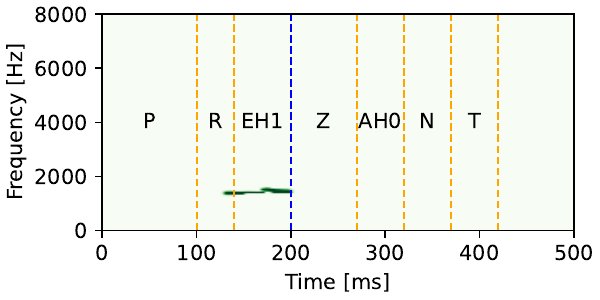}{6.0cm}{(b) F2 Heatmap}\hfill
    \rightfig{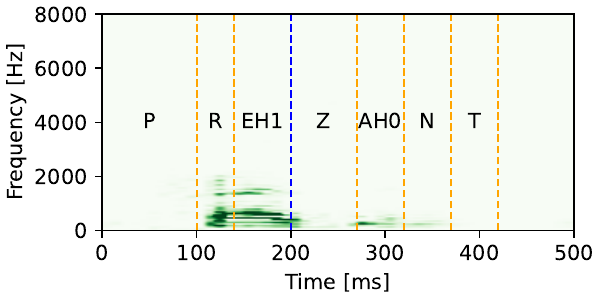}{6.0cm}{(c) VGG16+$\mathbf{LRP_{CMP}}$ Heatmap}
    }
    \caption{Heatmaps of the first two formants ($F_1$ and $F_2$) and the original heatmap generated by $\mathbf{LRP_{CMP}}$ using VGG16 for the word \emph{"PREsent"}. Panels (a) and (b) display the heatmaps generated for the two formants, considering bandwidth, spectrogram intensity, and voice intensity within the relevant time points. Finally, panel (c) illustrates the heatmap attributes generated by $\mathbf{LRP_{CMP}}$ utilizing VGG16. Orange vertical lines represent phoneme boundaries, and the blue vertical line represents the end of the initial syllable, same as Fig. \ref{fig:fig2}.}
    \label{fig:fig4}
\end{figure*}

\subsection{\label{subsec:6:4} Bounding Box and IOU Analysis}
The IOU measurements provide a quantitative assessment of the overlap between the bounding boxes and the relevant pixels in the LRP heatmaps. 

Table \ref{table:tab3} shows IOU measurements (\begin{math}\mu\end{math}) for different LRP methods, examining VGG16 and ResNet18. 
The $\mathbf{LRP_{CMP}}$, $\mathbf{LRP_{\alpha1}}$, $\mathbf{LRP_{\epsilon}}$ methods suggest that both classifiers attend primarily to acoustic information in stressed versus unstressed syllables, and that the greatest concentration of pixels is specifically in the stressed vowel region. The basic $\mathbf{LRP_z}$ method results are mostly consistent with these generalizations, but give some greater weight to unstressed syllables. 

Overall, these findings suggest that the classifiers predict the location of stress by attending to features of the stressed syllable, in particular to stressed vowels. However, it is important to note that under all methods, a considerable number of pixels lie outside of the stressed vowel -- consistent with the distribution of cues to lexical stress throughout the entire syllable. Furthermore, many pixels lie outside of the stressed syllable -- consistent with a view that sees lexical stress as a ''relational'' property of stressed as compared to unstressed syllables.

\begin{table}[ht]
    \centering
    \begin{ruledtabular}
    \caption{Average IOU values for different classifiers and LRP methods, separated by  stressed versus unstressed syllables and syllable region (vowel versus all other components of syllable).}
    \label{table:tab3}
    \begin{tabular}{l|l|cc|cc}      
        \multirow{2}{*}{Classifier} & \multirow{2}{*}{LRP Method} 
        & \multicolumn{2}{c|}{\textbf{$\mu_{\mathrm{vowel}}$}} 
        & \multicolumn{2}{c}{\textbf{$\mu_\mathrm{not\,vowel}$}} \\  
        & & Stress & No stress & Stress & No stress \\ 
        \hline
        \multirow{4}{*}{\vspace{0cm}\centering VGG16}
        & $\mathbf{LRP_{CMP}}$      & 0.432 & 0.135 & 0.276 & 0.156 \\
        & $\mathbf{LRP_{\alpha1}}$  & 0.419 & 0.129 & 0.279 & 0.173 \\
       & $\mathbf{LRP_{\epsilon}}$ & 0.428 & 0.137 & 0.27 & 0.164 \\
       & $\mathbf{LRP_z}$          & 0.301 & 0.154 & 0.318 & 0.226 \\
    \hline
       \multirow{4}{*}{\vspace{0cm}\centering ResNet18~~}
        & $\mathbf{LRP_{CMP}}$          & 0.37 & 0.136 & 0.298 & 0.195 \\
        & $\mathbf{LRP_{\alpha1}}$      & 0.361 & 0.128 & 0.255 & 0.255 \\
        & $\mathbf{LRP_{\epsilon}}$     &  0.34 & 0.13 & 0.275 & 0.254 \\
        & $\mathbf{LRP_z}$              & 0.326 & 0.12 & 0.204 & 0.364 \\
    \end{tabular}
    \end{ruledtabular}
\end{table}

\subsection{\label{subsec:6:6} Feature-Specific Relevance Analysis}

We extended our IOU analysis by investigating how VGG16's predictions are influenced by the phonetic features of the signal region that most strongly drives its responses, namely, the stressed vowels. We compared each of these features to the $\mathbf{LRP_{CMP}}$-derived heatmap, focusing on the heatmap region corresponding to the stressed vowel. Figure \ref{fig:fig4} shows two feature-specific heatmaps corresponding to $F_1$ and $F_2$ in isolation. Additionally, the original relevance heatmap produced by VGG16 and $\mathbf{LRP_{CMP}}$ is presented for comparison.

As shown in Table \ref{table:tab4}, a heatmap generated solely using $F_1$ of the stressed vowel exhibits the strongest correlation with the observed heatmap. Unsurprisingly, this formant makes a significant contribution, as it correlates with tongue height, jaw height, and the overall opening of the oral cavity, which in turn are strongly associated with vowel sound intensity. However, the combination of $F_1$ and $F_2$ is a close second to $F_1$ alone, suggesting that the two acoustic features traditionally related to vowel quality in English work together to determine classifier predictions. Similarly, heatmaps combining these two formants with $F_3$ and, to a lesser extent, $F_0$, are used to predict LRP responses. However, the outsize influence of $F_1$ is made clear by the low performance of the other formants and pitch in isolation. When additional features that are less relevant to the model’s actual decision-making are combined (e.g., $F_1$ + $F_2$ + $F_3$), the overall similarity with the LRP-derived heatmap tends to decrease. This reduction arises because the inclusion of non-informative or weakly correlated regions introduces additional variance, thereby lowering the $r$ value. In this sense, “more” does not necessarily mean “better”: although these additional features may still play supporting roles in classification, their weaker correspondence to the model’s main decision cues can dilute the overall correlation with the relevance pattern.

\begin{table}[ht]
    \centering
    \begin{ruledtabular}   
    \caption{$r$ and $p$ values for heatmaps based on different features in the stressed vowel region (averaged across samples), using VGG16 and applying \begin{math} \mathbf{LRP_{CMP}}\end{math} method, sorted by $r$. See text for details.}
    \label{table:tab4}
    \begin{tabular}{c|lcc}
        Method & Acoustic Feature & Mean $r\uparrow$ & Mean $p\downarrow$\\
        \hline
        \multirow{7}{*}{\vspace{-5.5cm}\centering \rothead{VGG16~~+~~$\mathbf{LRP_{CMP}}$}}
        & $F_1$ & 0.366 & 0.003\\
        & $F_1, F_2$ & 0.359 & 0.003\\
        & $F_1, F_2, F_3$ & 0.305 & 0.002\\
        & $F_0,F_1$ & 0.3 & 0.002\\
        & $F_1,F_3$ & 0.277 & 0.001\\        
        & $F_0,F_1,F_2$ & 0.245 & 0.001\\
        & $F_0,F_1,F_2,F_3$ & 0.237 & 0.001\\
        & $F_0,F_1,F_3$ & 0.205 & 0.001\\   
        & $F_0,F_2$ & 0.154 & 0.003\\  
        & $F_2$ & 0.154 & 0.008\\
        & $F_2,F_3$ & 0.136 & 0.008\\
        & $F_0$ & 0.116 & 0.004\\
        & $F_0, F_3$ & 0.114 & 0.003\\
        & $F_3$ & 0.048 & 0.02\\
    \end{tabular}
    \end{ruledtabular}    
\end{table}

While the $r$ values are significant, they suggest that these features, even when considered in combination, fail to explain many aspects of the LRP heatmap for stressed vowels. To explore this, we examine the frequency distribution of residual (unexplained) pixels. As shown in Table \ref{table:tab5}, the vast majority of unaccounted-for pixels lie in between $F_0$ and $F_1$.

What other acoustic features might explain these pixels? Resolving questions like this is a key area for development of this work. The phonetics literature suggests a number of possibilities (which are not mutually exclusive). Intensity has been proposed as a cue to stress \cite{cutler,gay1978physiological,gordon2017acoustic, vanHeuvenStress}. These pixels could reflect sensitivity to the concentration of energy in low frequencies during vowels.  A number of prior studies have identified spectral tilt / spectral balance as an important cue to lexical stress \cite{gordon2017acoustic,vanHeuvenStress}. Although spectral tilt measures often rely on information across a wider range of frequencies, spectral information in this range would provide a great deal of information about the steepness of spectral tilt. It is also possible that these pixels reflect sensitivity to harmonic structures. Inspection of Figures \ref{fig:fig2} and \ref{fig:fig3} suggests that the classifier is attending to regularly spaced concentrations of energy in these lower frequencies -- which could correspond to harmonics. The classifiers are using this information to track pitch, a contributor to lexical stress perception \cite{cutler}. Our classifier relies on a standard spectrographic representation as input. The relatively poor resolution of this representation at low frequencies makes it a poor estimator of pitch. Tracking the spacing of harmonics (which have high amplitude in this region of the signal) would give the classifier access to this critical information. This could allow the classifier to estimate pitch with a higher degree of resolution than is available in the region of the spectrogram corresponding to $F_0$, using information that can be readily uncovered from the spectrographic input. It is also possible that the classifier is using the first and second harmonic amplitude difference, H1-H2, to measure voice quality \cite{garellek2019phonetics} differences related to stress \cite{gordon2017acoustic}.

\begin{table}[ht]
    \centering
    \begin{ruledtabular}
    \caption{Relative distribution of pixels (averaged across samples) in the stressed vowel region of the residual \begin{math}\mathbf{LRP_{CMP}}\end{math} heatmap for VGG16 (see text for details).}
    \label{table:tab5}
    \begin{tabular}{l|cc}
        Region & Initial stressed Vowel & Final stressed Vowel\\
        \hline
        \raggedright Between ${F_0,F_1}$ & 0.93 & 0.886\\
        \raggedright Between ${F_1,F_2}$ & 0.07 & 0.1134\\
        \raggedright Between ${F_2,F_3}$ & 0 & 0\\
        \raggedright Above ${F_3}$ & 0 & 0\\
    \end{tabular}
    \end{ruledtabular}
\end{table}

\begin{figure}[htbp]
    \centering
   \includegraphics[scale=0.5]{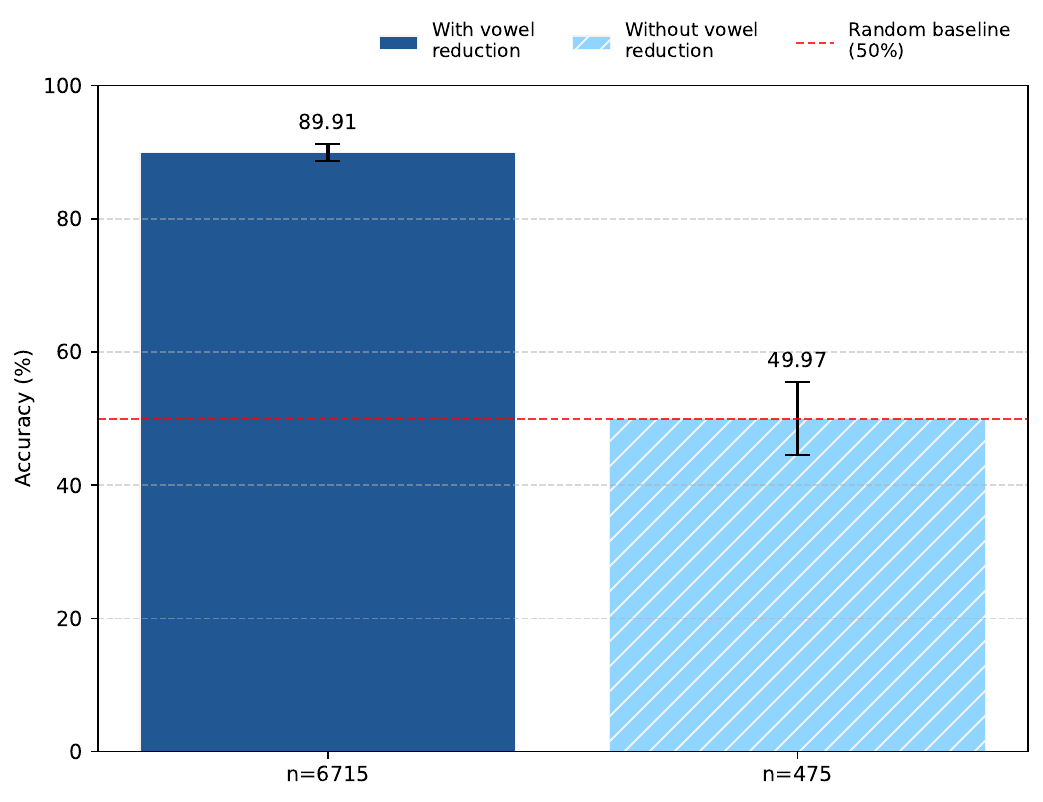}
    \caption{Classification accuracy of the VGG16 model on the minimal pairs test set, aggregated by vowel reduction in the unstressed syllable. Bars show mean per-word accuracy, averaged across minimal pairs with vowel reduction (left) and without vowel reduction (right). Error bars indicate \(\pm\) standard error of the mean computed across words within each group. The number of test samples in each group is indicated by n. Dashed red horizontal line denotes chance-level (random baseline) accuracy at 50\%.}
    \label{fig:fig5}
\end{figure}

\section{\label{sec:7} Discussion}
To explore what allows CNNs to outperform more traditional approaches, we built several novel classifiers for lexical stress detection in English disyllabic words, using LRP-based analyses to explore the spatiotemporal properties that drive model performance. A fully automatic pipeline was used to extract disyllabic words from read and spontaneous speech. CNNs trained on words without minimal pairs exhibited high performance on held-out no minimal pair words and words with minimal pairs. LRP analyses on the spectrogram input space for the CNNs revealed that the models were sensitive to acoustic properties in both stressed and unstressed syllables, but showed greater reliance on stressed syllables (particularly the stressed vowel). Analysis of our highest-performing model showed that $F_1$ of the stressed vowel exerted a stronger influence on predictions than other formants, but $F_2$ and, to a lesser extent, $F_3$ also contributed. Furthermore, these features failed to capture a substantial amount of variance in the LRP heatmap, suggesting additional features of the stressed vowel contribute to its predictions (e.g., spectral balance and pitch based on spacing of harmonics).

The classifiers' greater reliance on the properties of stressed versus unstressed syllables and more specifically stressed vowels versus other syllable regions is similar to the focus of the phonetics literature on stressed syllables and stressed vowels \cite{cutler, gay1978physiological}. Critically, differing from some early phonetics studies, the classifier does not focus \emph{exclusively} on these regions, but also attends to information in both syllables, including non-vowel portions \footnote{It is possible that the distributed nature of stress cues reflects errors in our forced aligner. However, given the substantial temporal range of LRP pixels (see, e.g., Figure \ref{fig:fig4}(c)), it's likely that the results are robust to small changes to segment boundaries.}. 
The use of information in the unstressed syllable resonates with the relatively widespread use of relativized or normalized cues to lexical stress \cite{cutler,vanHeuvenStress}. The use of information outside of the vowel is also consistent with phonetics research that examines duration and intensity of the whole syllable, the rime, or consonants  \cite{cutler, gordon2017acoustic}. More broadly, the use of consonantal information is consistent with the view that stress is not a property of vowels but of whole syllables, organized within more complex prosodic structures such as feet; these structures condition not only the properties of vowels but also surrounding consonants in stressed and unstressed syllables (see, e.g., \cite{jensen2000against}). Future work should continue to develop this general line of analysis by examining variation in stress driven by larger sentential and discourse contexts (see \citet{vanHeuvenStress} for a review and discussion).

While, in English, stressed versus unstressed syllables are typically distinguished by full versus reduced vowel quality, some disyllabic stress-pairs are pronounced with full vowels in both syllables. Figure \ref{fig:fig5} summarizes classification accuracy on the minimal pairs test set after aggregating all minimal-pair words into two groups based on the presence or absence of vowel reduction in the unstressed syllable. Because the task is binary (initial vs. final stress), chance-level performance corresponds to 50\% accuracy.
Tagging of minimal pair words as ''with'' or ''without'' vowel reduction was based on a predefined list of English stress pairs \citep{bosker_lexical_stress_minimal_pairs_2025}. Words treated as lacking vowel reduction include \emph{import}, \emph{insult}, \emph{digest}, \emph{increase}, and \emph{transfer}. This classification was not based on token-level acoustic annotation; consequently, the actual degree of reduction in particular productions may not precisely align with these assumptions.
 The figure reveals a clear performance gap: the models achieve substantially higher accuracy for words exhibiting vowel reduction than for those that do not. When vowel reduction is absent, mean accuracy falls to chance level.
This pattern is consistent with the nature of the training data, which consisted primarily of no minimal pairs disyllabic words in which vowel reduction serves as an important acoustic correlate of stress \cite{vanHeuvenStress}. It also supports the interpretation that formant information, particularly cues related to vowel reduction, plays a central role in model performance, while also confirming that the models are sensitive to additional cues in cases where vowel quality differences are minimal.

Our feature-specific relevance analysis allowed us to connect the LRP results to formants of the stressed vowel. However, these analyses left a great deal of the LRP signal unexplained -- not only in the stressed vowel, but also in other regions of the stressed syllable and the entirety of the unstressed syllable. This is consistent with prior work suggesting that formant information alone is insufficient for lexical stress classification \cite{bentum2024processing}. Future work examining what acoustic features these pixels correspond to may provide new insights into the phonetics of English stress.

More broadly, this work suggests that deep learning may offer an important complement to traditional approaches to phonetics. Rather than starting with highly controlled materials like minimal pairs elicited under laboratory conditions, our classifier was trained on ''messy'' data -- uncontrolled materials, varying along many dimensions beyond the contrast of interest, drawn from speech produced by a number of talkers in a variety of conditions. When exposed to this (naturally occurring) variability, the classifiers were able to extract regularities that fit with the previous phonetics literature. With increasingly powerful interpretability tools, deep learning may help phoneticians better understand the acoustic properties associated with linguistic contrasts.

\section*{Acknowledgments}
Supported by NSF DRL Grant No. 2219843 and BSF Grant No. 2022618.

\section*{Author Declarations}
\subsection*{Conflict of Interests}
The authors have no conflict of interests to declare.

\section*{Data Availability}
The dataset used in this study is publicly available \citep{mlspeech_lexical_stress_2025}. Code and a demo are also publicly available \citep{allouche_minimal_pairs_lexical_stress}.

\bibliography{lexical_stress_lrp.bib}

@incollection{garellek2019phonetics,
  title={The phonetics of voice 1},
  author={Garellek, Marc},
  booktitle={The Routledge handbook of phonetics},
  pages={75--106},
  year={2019},
  publisher={Routledge}
}

@inproceedings{bentum2024processing,
  title={The processing of stress in end-to-end automatic speech recognition models},
  author={Bentum, Martijn and ten Bosch, Louis and Lentz, Tom},
  booktitle={Proc. Interspeech},
  volume={2024},
  pages={2350--2354},
  year={2024}
}

@inproceedings{vanHeuvenStress,
  title={Acoustic correlates and perceptual cues of word and sentence stress},
  author={Van Heuven, Vincent J},
  booktitle={The study of word stress and accent: Theories, methods and data},
  pages={15--59},
  year={2018},
  publisher={Cambridge Univ. Press Cambridge, UK}
}

@article{gordon2017acoustic,
  title={Acoustic correlates of word stress: A cross-linguistic survey},
  author={Gordon, Matthew and Roettger, Timo},
  journal={Linguistics Vanguard},
  volume={3},
  number={1},
  pages={20170007},
  year={2017},
  publisher={De Gruyter}
}

@article{fry1955duration,
  title={Duration and intensity as physical correlates of linguistic stress},
  author={Fry, Dennis B},
  journal={The Journal of the Acoustical Society of America},
  volume={27},
  number={4},
  pages={765--768},
  year={1955},
  publisher={AIP Publishing}
}

@article{fry1958experiments,
  title={Experiments in the perception of stress},
  author={Fry, Dennis B},
  journal={Language and speech},
  volume={1},
  number={2},
  pages={126--152},
  year={1958},
  publisher={SAGE Publications Sage UK: London, England}
}

@article{lieberman1960some,
  title={Some acoustic correlates of word stress in American English},
  author={Lieberman, Philip},
  journal={The Journal of the Acoustical Society of America},
  volume={32},
  number={4},
  pages={451--454},
  year={1960},
  publisher={Acoustical Society of America}
}

@InProceedings{1415269,
  author =    "Tepperman, J. and Narayanan, S.",
  title =     "Automatic syllable stress detection using prosodic features for pronunciation evaluation of language learners",
  booktitle = "Proceedings. (ICASSP '05). IEEE International Conference on Acoustics, Speech, and Signal Processing, 2005.",
  year =      "2005",
  volume =    "1",
  number =    "",
  pages =     "I/937-I/940 Vol. 1",
  doi =       "10.1109/ICASSP.2005.1415269"
}

@INPROCEEDINGS{1168788,
  author={Waibel, A.},
  booktitle={ICASSP '86. IEEE International Conference on Acoustics, Speech, and Signal Processing}, 
  title={Recognition of lexical stress in a continuous speech understanding system - A pattern recognition approach}, 
  year={1986},
  volume={11},
  number={},
  pages={2287-2290},
  doi={10.1109/ICASSP.1986.1168788}}

@INPROCEEDINGS{607932,
  author={Ying, G.S. and Jamieson, L.H. and Ruxin Chen and Michell, C.D. and Hsin Liu},
  booktitle={Proceeding of Fourth International Conference on Spoken Language Processing. ICSLP '96}, 
  title={Lexical stress detection on stress-minimal word pairs}, 
  year={1996},
  volume={3},
  number={},
  pages={1612-1615 vol.3},
  doi={10.1109/ICSLP.1996.607932}}

@inproceedings{shahin2016automatic,
  title={Automatic Classification of Lexical Stress in English and Arabic Languages Using Deep Learning.},
  author={Shahin, Mostafa Ali and Epps, Julien and Ahmed, Beena},
  booktitle={Interspeech},
  pages={175--179},
  year={2016}
}

@article{LI201828,
title = {Automatic lexical stress and pitch accent detection for L2 English speech using multi-distribution deep neural networks},
journal = {Speech Communication},
volume = {96},
pages = {28-36},
year = {2018},
issn = {0167-6393},
doi = {https://doi.org/10.1016/j.specom.2017.11.003},
url = {https://www.sciencedirect.com/science/article/pii/S0167639315300637},
author = {Kun Li and Shaoguang Mao and Xu Li and Zhiyong Wu and Helen Meng}
}

@article{korzekwa2020detection,
  title={Detection of lexical stress errors in non-native (l2) english with data augmentation and attention},
  author={Korzekwa, Daniel and Barra-Chicote, Roberto and Zaporowski, Szymon and Beringer, Grzegorz and Lorenzo-Trueba, Jaime and Serafinowicz, Alicja and Droppo, Jasha and Drugman, Thomas and Kostek, Bozena},
  journal={arXiv preprint arXiv:2012.14788},
  year={2020}
}

@INPROCEEDINGS{7178964,
  author={Panayotov, Vassil and Chen, Guoguo and Povey, Daniel and Khudanpur, Sanjeev},
  booktitle={2015 IEEE International Conference on Acoustics, Speech and Signal Processing (ICASSP)}, 
  title={Librispeech: An ASR corpus based on public domain audio books}, 
  year={2015},
  volume={},
  number={},
  pages={5206-5210},
  doi={10.1109/ICASSP.2015.7178964}}

@article{spaeth2014supreme,
  title={Supreme Court database code book},
  author={Spaeth, Harold and Epstein, Lee and Ruger, Ted and Whittington, Keith and Segal, Jeffrey and Martin, Andrew D},
  journal={URL: http://scdb. wustl. edu},
  year={2014}
}

@inproceedings{hernandez2018ted,
  title={TED-LIUM 3: Twice as much data and corpus repartition for experiments on speaker adaptation},
  author={Hernandez, Fran{\c{c}}ois and Nguyen, Vincent and Ghannay, Sahar and Tomashenko, Natalia and Esteve, Yannick},
  booktitle={Speech and Computer: 20th International Conference, SPECOM 2018, Leipzig, Germany, September 18--22, 2018, Proceedings 20},
  pages={198--208},
  year={2018},
  organization={Springer}
}

@inproceedings{mcauliffe2017montreal,
  title={Montreal forced aligner: Trainable text-speech alignment using kaldi.},
  author={McAuliffe, Michael and Socolof, Michaela and Mihuc, Sarah and Wagner, Michael and Sonderegger, Morgan},
  booktitle={Interspeech},
  volume={2017},
  pages={498--502},
  year={2017}
}

@article{simonyan2014very,
  title={Very deep convolutional networks for large-scale image recognition},
  author={Simonyan, Karen},
  journal={arXiv preprint arXiv:1409.1556},
  year={2014}
}

@ARTICLE{726791,
  author={Lecun, Y. and Bottou, L. and Bengio, Y. and Haffner, P.},
  journal={Proceedings of the IEEE}, 
  title={Gradient-based learning applied to document recognition}, 
  year={1998},
  volume={86},
  number={11},
  pages={2278-2324},
  doi={10.1109/5.726791}}

@article{bach2015pixel,
  title={On pixel-wise explanations for non-linear classifier decisions by layer-wise relevance propagation},
  author={Bach, Sebastian and Binder, Alexander and Montavon, Gr{\'e}goire and Klauschen, Frederick and M{\"u}ller, Klaus-Robert and Samek, Wojciech},
  journal={PloS one},
  volume={10},
  number={7},
  pages={e0130140},
  year={2015},
  publisher={Public Library of Science San Francisco, CA USA}
}

@InProceedings{Lapuschkin2016CVPR,
author = {Lapuschkin, Sebastian and Binder, Alexander and Montavon, Gregoire and Muller, Klaus-Robert and Samek, Wojciech},
title = {Analyzing Classifiers: Fisher Vectors and Deep Neural Networks},
booktitle = {Proceedings of the IEEE Conference on Computer Vision and Pattern Recognition (CVPR)},
month = {June},
year = {2016}
}

@article{Annex2020, doi = {10.21105/joss.02050}, url = {https://doi.org/10.21105/joss.02050}, year = {2020}, publisher = {The Open Journal}, volume = {5}, number = {46}, pages = {2050}, author = {Andrew M. Annex and Ben Pearson and Benoît Seignovert and Brian T. Carcich and Helge Eichhorn and Jesse A. Mapel and Johan L. Freiherr von Forstner and Jonathan McAuliffe and Jorge Diaz del Rio and Kristin L. Berry and K.-Michael Aye and Marcel Stefko and Miguel de Val-Borro and Shankar Kulumani and Shin-ya Murakami}, title = {SpiceyPy: a Pythonic Wrapper for the SPICE Toolkit}, journal = {Journal of Open Source Software} }

@article{kingma2014adam,
  title={Adam: A method for stochastic optimization},
  author={Kingma, Diederik P},
  journal={arXiv preprint arXiv:1412.6980},
  year={2014}
}

@INPROCEEDINGS{9206975,
  author={Kohlbrenner, Maximilian and Bauer, Alexander and Nakajima, Shinichi and Binder, Alexander and Samek, Wojciech and Lapuschkin, Sebastian},
  booktitle={2020 International Joint Conference on Neural Networks (IJCNN)}, 
  title={Towards Best Practice in Explaining Neural Network Decisions with LRP}, 
  year={2020},
  volume={},
  number={},
  pages={1-7},
  doi={10.1109/IJCNN48605.2020.9206975}}

@inproceedings{lin2017focal,
  title={Focal loss for dense object detection},
  author={Lin, Tsung-Yi and Goyal, Priya and Girshick, Ross and He, Kaiming and Doll{\'a}r, Piotr},
  booktitle={Proceedings of the IEEE international conference on computer vision},
  pages={2980--2988},
  year={2017}
}

@InProceedings{He2016CVPR,
author = {He, Kaiming and Zhang, Xiangyu and Ren, Shaoqing and Sun, Jian},
title = {Deep Residual Learning for Image Recognition},
booktitle = {Proceedings of the IEEE Conference on Computer Vision and Pattern Recognition (CVPR)},
month = {June},
year = {2016}
}

@article{pratt,
author = "Boersma, Paul and Weenink, David",
title = "Praat: doing phonetics by computer",
version = "6.2.09",
date = "2022-02-15",
url = "https://www.praat.org/",
urldate = "2025-10-15",
note = "Computer program"
}

@article{hurst2024gpt,
  title={Gpt-4o system card},
  author={Hurst, Aaron and Lerer, Adam and Goucher, Adam P and Perelman, Adam and Ramesh, Aditya and Clark, Aidan and Ostrow, AJ and Welihinda, Akila and Hayes, Alan and Radford, Alec and others},
  journal={arXiv preprint arXiv:2410.21276},
  year={2024}
}

@incollection{cutler,
  author={Cutler, Anne},
  year= {2005}, 
  title={Lexical stress}, 
  editor = {D. B. Pisoni and R. E. Remez}, 
  booktitle= {The Handbook of Speech Perception}, 
  publisher= {Blackwell},
  address= {Oxford},
  pages={264--289}
}

@article{yu2010cross,
  title={A cross-language study of perception of lexical stress in English},
  author={Yu, Vickie Y and Andruski, Jean E},
  journal={Journal of psycholinguistic research},
  volume={39},
  pages={323--344},
  year={2010},
  publisher={Springer}
}

@inproceedings{richey2018voices,
  title={Voices obscured in complex environmental settings (voices) corpus},
  author={Richey, Colleen and Barrios, Maria A and Armstrong, Zeb and Bartels, Chris and Franco, Horacio and Graciarena, Martin and Lawson, Aaron and Nandwana, Mahesh Kumar and Stauffer, Allen and van Hout, Julien and others},
  booktitle={Interspeech},
  year={2018},
  pages={1566--1570}
}

@article{gay1978physiological,
  title={Physiological and acoustic correlates of perceived stress},
  author={Gay, Thomas},
  journal={Language and Speech},
  volume={21},
  number={4},
  pages={347--353},
  year={1978},
  publisher={SAGE Publications Sage UK: London, England}
}

@article{jensen2000against,
  title={Against ambisyllabicity},
  author={Jensen, John T},
  journal={Phonology},
  volume={17},
  number={2},
  pages={187--235},
  year={2000},
  publisher={Cambridge University Press}
}

@article{bain2023whisperx,
  title={Whisperx: Time-accurate speech transcription of long-form audio},
  author={Bain, Max and Huh, Jaesung and Han, Tengda and Zisserman, Andrew},
  journal={arXiv preprint arXiv:2303.00747},
  year={2023}
}

@article{pratap2024scaling,
  title={Scaling speech technology to 1,000+ languages},
  author={Pratap, Vineel and Tjandra, Andros and Shi, Bowen and Tomasello, Paden and Babu, Arun and Kundu, Sayani and Elkahky, Ali and Ni, Zhaoheng and Vyas, Apoorv and Fazel-Zarandi, Maryam and others},
  journal={Journal of Machine Learning Research},
  volume={25},
  number={97},
  pages={1--52},
  year={2024}
}

@article{rousso2024tradition,
  title={Tradition or innovation: A comparison of modern asr methods for forced alignment},
  author={Rousso, Rotem and Cohen, Eyal and Keshet, Joseph and Chodroff, Eleanor},
  journal={arXiv preprint arXiv:2406.19363},
  year={2024}
}

@dataset{mlspeech_lexical_stress_2025,
  author={{MLSpeech}},
  title={Lexical Stress Dataset},
  year={2025},
  publisher={Hugging Face},
  url={https://huggingface.co/datasets/MLSpeech/lexical_stress_dataset},
}

@software{allouche_minimal_pairs_lexical_stress,
  author={Allouche, Itai},
  title={Minimal Pairs Lexical Stress},
  year={2025},
  url={https://github.com/ItaiAllouche/minimalPairsLexicalStress},
  note={GitHub repository}
}

@software{p2tk_syllabifier_2014,
  author={{P2TK Developers}},
  title={P2TK: Phonetic and Phonological Toolkit},
  year={2014},
  url={https://sourceforge.net/p/p2tk/},
  note={Syllabified dictionary and syllabifier. Accessed: 2026-01-04}
}

@software{kokhlikyan_captum_2020,
  author={Kokhlikyan, Narine and Miglani, Vivek and Martin, Michael and Wang, Edward and Alsallakh, Bilal and Reynolds, Jonathan and Melnikov, Alexander and Kliushkina, Natalia and Araya, Carlos and Yan, Siqi and Reblitz-Richardson, Thomas},
  title={Captum: A Unified and Generic Model Interpretability Library for PyTorch},
  year={2020},
  url={https://captum.ai},
  note={Computer software. Accessed: 2026-01-04}
}

@dataset{bosker_lexical_stress_minimal_pairs_2025,
  author={Bosker, Hans Rutger},
  title={List of minimal pairs differing only in lexical stress},
  year={2024},
  publisher={OSF},
  doi={10.17605/OSF.IO/5D4KS},
  url={https://osf.io/5d4ks/},
  note={Accessed: 2026-01-04}
}

\end{document}